\documentclass[11pt]{article}
\usepackage{eamt23}
\usepackage{times}
\usepackage{url}
\usepackage{latexsym}
\usepackage[small,bf]{caption} 
\usepackage[utf8]{inputenc}
\usepackage{booktabs}
\usepackage{graphicx}
\usepackage{amsmath}
\usepackage{amsfonts}
\usepackage{subcaption}

\usepackage{float}
\usepackage{todonotes}
\usepackage{textcomp}
\usepackage{listings}
\usepackage{csquotes}

\usepackage{tabularx}
\usepackage{multirow}

\usepackage{multirow}
\usepackage{siunitx}

\usepackage{xcolor}
\newcommand{\hl}[1]{\textcolor{red}{\underline{#1}}}

\title{Enhancing Supervised Learning with Contrastive Markings\\ in Neural Machine Translation Training}

\author{Nathaniel Berger$^{\ast}$, Miriam Exel$^\ddag$, Matthias Huck$^\ddag$ \and Stefan Riezler$^{\dagger,\ast}$\\ 
  $^{\ast}$Computational Linguistics \& $^\dagger$IWR, Heidelberg University, Germany \\
  $^\ddag$SAP SE, Dietmar-Hopp-Allee 16, 69190 Walldorf, Germany\\
  {\tt \{berger, riezler\}@cl.uni-heidelberg.de}\\  
  {\tt \{miriam.exel, matthias.huck\}@sap.com}}

\date{}

\begin{document}
\maketitle
\begin{abstract}
Supervised learning in Neural Machine Translation (NMT) typically follows a teacher forcing paradigm where reference tokens constitute the conditioning context in the model's prediction, instead of its own previous predictions. In order to alleviate this lack of exploration in the space of translations, we present a simple extension of standard maximum likelihood estimation by a contrastive marking objective. The additional training signals are extracted automatically from reference translations by comparing the system hypothesis against the reference, and used for up/down-weighting correct/incorrect tokens. The proposed new training procedure requires one additional translation pass over the training set per epoch, and does not alter the standard inference setup. We show that training with contrastive markings yields improvements on top of supervised learning, and is  especially useful when learning from postedits where contrastive markings indicate human error corrections to the original hypotheses. Code is publicly released\footnote{\url{https://www.cl.uni-heidelberg.de/statnlpgroup/contrastive_markings/}}.
\end{abstract}

\section{Introduction}

Due to the availability of large parallel data sets for most language pairs, the standard training procedure in Neural Machine Translation (NMT) is supervised learning of a maximum likelihood objective where reference tokens constitute the target history in the conditional language model, instead of the model's own predictions. Feeding back the reference history in model training, known as \emph{teacher forcing} \cite{WilliamsZipser:89}, encourages the sequence model to stay close to the reference sequence, but prevents the model to learn how to predict conditioned on its own history, which is the actual task at inference time. This lack of exploration in learning has been dubbed \emph{exposure bias} by \newcite{RanzatoETAL:16}. It has been tackled by techniques that explicitly integrate the model's own prediction history into training, e.g.\,scheduled sampling \cite{BengioETAL:15}, minimum risk training \cite{ShenETAL:16}, reinforcement learning \cite{BahdanauETAL:17}, imitation learning \cite{lin-etal-2020-autoregressive}, or ramp loss \cite{jehl-etal-2019-learning}, amongst others. In most of these approaches, feedback from a human expert is simulated by comparing a system translation against a human reference according to an automatic evaluation metric, and by extracting a sequence- or token-level reward signal from the evaluation score. 

In this paper, we present a method to incorporate \emph{contrastive markings} of differences between the model's own predictions and references into the learning objective. Our approach builds on previous work on integrating weak human feedback in form of error markings as supervision signal in NMT training \cite{kreutzer-etal-2020-correct}. This work was conceptualized for reducing human annotation effort in interactive machine translation, however, it can also be used on simulated error markings extracted from an automatic evaluation score. It allows the model to extract a contrastive signal from the reference translation that can be used to reinforce or penalize correct or incorrect tokens in the model's own predictions.  Such a reward signal is more fine-grained than a sequence-level reward obtained by a sequence-level automatic evaluation metric, and less noisy than token-based rewards obtained by reward shaping \cite{NgETAL:99}. 

Our hypothesis is that such contrastive markings should be especially useful in learning setups where human postedits are used as reference signals. In such scenarios, contrastive markings are likely to indicate erroneous deviations of machine translations from human error corrections, instead of penalizing correct translations that happen to deviate from independently constructed human reference translations.
We confirm this hypothesis by simulating a legacy machine translation system for which human postedits are available by performing knowledge distillation \cite{kim-rush-2016-sequence} on the stored legacy machine translations. We define a ``legacy'' machine translation system as a system which was previously used in production and produced translations for which human feedback was gathered, but which is no longer productive. Knowledge distillation is required because the legacy system is a black-box system that is unavailable to us, but its outputs are available. For comparison, we apply our framework to standard parallel data where reference translations were generated from scratch. Our experimental results show that on both datasets, combining teacher forcing on post\-edits with learning from error markings, improves results with respect to TER on test data, with larger improvements for the knowledge-distilled model that emulates outputs of the legacy system.

A further novelty of our approach is the true online learning setup where new error markings are computed after every epoch of model training, instead of using constant simulated markings that are pre-computed from fixed machine translation outputs as in previous work \cite{PetrushkovETAL:18,grangier-auli-2018-quickedit,kreutzer-etal-2020-correct}. Online error markings can be computed in a light-weight fashion by longest common subsequence calculations.
The overhead incurred by the new training procedure is one additional translation pass over the training set, whereas at inference time the system does not require additional information, but can be shown to produce improved translations based on the proposed improved training setup.



\section{Related Work}





Most approaches to remedy the exposure bias problem simulate a sentence-level reward or cost function from an automatic evaluation metric and incorporate it into a reinforcement- or imitation-learning setup \cite{RanzatoETAL:16,ShenETAL:16,BahdanauETAL:17,lin-etal-2020-autoregressive,jehl-etal-2019-learning,gu2019levenshtein,xu-carpuat-2021-editor}.

Methods that are conceptualized to work directly with human postedits integrate the human feedback signal more directly, without the middleman of an automatic evaluation heuristic. The standard learning paradigm is supervised learning where postedits are treated as reference translations (see, for example, \newcite{TurchiETAL:17}). Most approaches to learning from error markings adapt the supervised learning objective to learn from correct tokens in partial translations
\cite{MarieMax:15,PetrushkovETAL:18,DomingoETAL:17,kreutzer-etal-2020-correct}. 


The QuickEdit \cite{grangier-auli-2018-quickedit} approach uses the hypothesis produced by an NMT system and token-level markings as an extra input to an automatic postediting system (APE), and additionally requires markings on the system output at inference time. This requires a dual encoder architecture with the decoder attending to both the source and hypothesis encoders. In this case,  convolutional encoders and decoders of \newcite{GehringETAL:17} are used. 

Our approach builds upon the work of \newcite{PetrushkovETAL:18} and \newcite{kreutzer-etal-2020-correct} who incorporate token-level markings as learning signal into NMT training. In contrast to \newcite{grangier-auli-2018-quickedit}, who compute markings off\-line before training and require them for inference, we only require them during training and calculate markings on\-line. Furthermore, instead of presenting markings to the system as an extra input, they are integrated into the objective function as a weight. While \newcite{PetrushkovETAL:18} simulate markings from reference translations by extracting deletion operations from longest common subsequence calculations, \newcite{kreutzer-etal-2020-correct} show how to learn from markings solicited from human annotators. In contrast to these approaches, we integrate markings to enhance supervised learning in a true online fashion.

\begin{table*}[t!]
    \centering
    \begin{tabularx}{\linewidth}{l X}
    \toprule
        Source & To remove the highlighting , un@@ mark the menu entry . \\
         \hline
         Hypothesis & Um die Her@@ vor@@ hebung \hl{zu entfernen} , \hl{mark@@ ieren} Sie \hl{den} Menü@@ ein@@ trag . \\
         \hline
         Reference & Um die Her@@ vor@@ hebung auszu@@ schalten , de@@ aktivieren Sie diesen Menü@@ ein@@ trag .\\
         \hline
         Markings & 1 1 1 1 1 \hl{0 0} 1 \hl{0 0} 1 \hl{0} 1 1 1 1\\ 
    \bottomrule
    \end{tabularx}
    \caption{An example of a source, hypothesis, and reference triple along with the contrastive markings generated by comparing the hypothesis to the reference. Markings of $1$ indicate a correct subword token, while $0$ indicates an incorrect subword token. We used byte-pair encoding \cite{sennrich-etal-2016-neural} and the "@@" indicate that this token is part of the same word as the following token. We underline and color the incorrect tokens and their corresponding markings red.
    }
    \label{tab:datapoint_example}
\end{table*}

\section{Methods}

\begin{figure*}[h!]
\centering
    \includegraphics{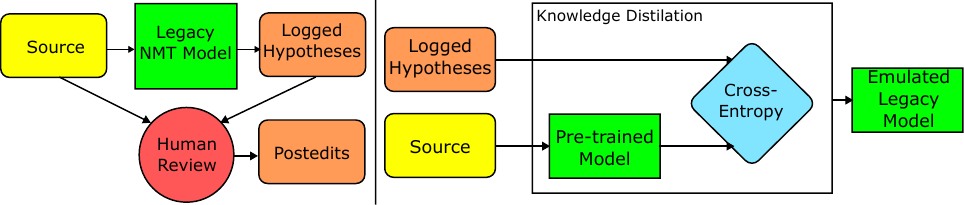}
    \caption{\textbf{Left:} The WMT21 APE dataset is created by having a black-box NMT system generate hypothesis translations. These logged hypotheses are then given to human reviewers to postedit to create a triple of (source, hypothesis, postedit). \\ \textbf{Right:} Because the system that generated the hypotheses is not available for us to fine-tune, we try to emulate it with knowledge distillation. We train the model to reproduce the original hypothesis by using them as targets with a cross-entropy loss to produce an emulated legacy model.}
    \label{fig:postedits_and_kd}

\end{figure*}

\subsection{Learning Objectives}

Let $x = x_1 \dots x_S$ be a sequence of indices over a source vocabulary $\mathcal{V}_{\textsc{Src}}$, and $y = y_1 \dots y_T$ a sequence of indices over a target vocabulary $\mathcal{V}_{\textsc{Trg}}$. The goal of sequence-to-sequence learning is to learn a function for mapping an input sequence $x$  into an output sequence $y$. For the example of machine translation, $y$ is a translation of $x$, and a model parameterized by a set of weights $\theta$ is optimized to maximize $p_{\theta}(y \mid x)$. This quantity is further factorized into conditional probabilities over single tokens 
$p_{\theta}(y \mid x) = \prod^{T}_{t=1}{p_{\theta}(y_t \mid x; y_{<t})}$, 
where the latter distribution is defined by the neural model's softmax-normalized output vector:
\begin{align}\label{eq:softmax}
    p_{\theta}(y_t \mid x; y_{<t}) = \texttt{softmax}(\text{NN}_{\theta}(x; y_{<t})).
\end{align}
There are various options for building the architecture of the neural model $\text{NN}_{\theta}$, such as recurrent \cite{BahdanauETAL:15}, convolutional \cite{GehringETAL:17} or attention-based \cite{vaswani-2017} encoder-decoder architectures.

Standard supervised learning from postedits treats a postedited output translation $y^*$ for an input $x$ the same as a human reference translation \cite{TurchiETAL:17} by maximizing the likelihood of the user-corrected outputs where 
\begin{align}\label{eq:loss}
    L_{PE}(\theta) = \sum_{x,y^*} \sum_{t=1}^T \log p_{\theta}(y^*_t \mid x; y^*_{<t}),
\end{align}
 using stochastic gradient descent techniques \cite{BottouETAL:18}.

\newcite{PetrushkovETAL:18} suggested learning from error markings $\delta^m_t$ of tokens $t$ in machine-generated output $\hat{y}$. Denote $\delta^+_t$ if marked as correct, or $\delta^-_t$ otherwise,  than a model with $\delta^+_t=1$ and $\delta^-_t=0$ will reward correct tokens and ignore incorrect outputs. 
The objective of the learning system is to maximize the likelihood of the correct parts of the output where 
\begin{align} \label{eq:weighted-loss}
  L_{M}(\theta) = \sum_{x,\hat{y}} \sum_{t=1}^T \delta^m_t \log p_{\theta}(\hat{y}_t \mid x; \hat{y}_{<t}).
\end{align}

The tokens $\hat{y}_t$ that receive $\delta_t=1$ are part of the correct output $y^{*}$, so the model receives a strong signal how a corrected output should look like. Although the likelihood of the incorrect parts of the sequence does not weigh into the sum, they are contained in the context of the correct parts (in $\hat{y}_{<t}$). Alternatively, it might be beneficial to penalize incorrect tokens, with e.g. $\delta^-_t=-0.5$, and reward correct tokens $\delta^+_t=0.5$, which aligns with the findings of \newcite{LamETAL:19}.

Our final combined objective is a linear interpolation of the log-likelihood of postedits $L_{PE}$ and the log-likelihood of markings $L_{M}$:
\begin{align}\label{eq:combined-loss}
    L(\theta) = \alpha L_{PE} + (1-\alpha) L_{M}.
\end{align}

\begin{table*}[t!]
    \centering
    \begin{tabular}{ccccccc}
    \toprule
    System & \multicolumn{2}{c}{Train} & \multicolumn{2}{c}{Dev} & \multicolumn{2}{c}{Test}\\
    & BLEU & TER & BLEU & TER & BLEU & TER\\
    \midrule
    APE MT Outputs & 100.0 & 0.0 & 100.0 & 0.0 & 100.0 & 0.0 \\
    \midrule
    Baseline Model & 48.0 & 31.8 & 49.0 & 31.0 & 46.2 & 33.8 \\
    KD Model & 88.9 & 5.8 & 56.0 & 25.9 & 55.8 & 26.7 \\
    \bottomrule
    \end{tabular}
    \caption{Systems outputs compared to APE data \emph{MT outputs}. BLEU and TER scores indicate distance of system outputs to MT outputs that were shown to human posteditors. Results show that Knowledge Distillation (KD) on APE MT Outputs improves distances (higher BLEU, lower TER), enabling improved approximation of the MT system that generated the hypotheses used in the APE dataset. Baseline and Knowledge Distillation systems evaluated with a beam size of 5.}
    \label{tab:ape_mt_output_results}

\end{table*}

\begin{table*}[t!]
    \centering
    \begin{tabular}{ccccccc}
    \toprule
    System & \multicolumn{2}{c}{Train} & \multicolumn{2}{c}{Dev} & \multicolumn{2}{c}{Test}\\
    & BLEU & TER & BLEU & TER & BLEU & TER\\
    \midrule
    APE MT Outputs & 70.8 & 18.1 & 69.1 & 18.9 & 71.5 & 17.9 \\
    \midrule
    Baseline Model & 42.4 & 36.9 & 43.3 & 35.8 & 41.7 & 37.8 \\
    KD Model & 66.0 & 20.8 & 49.1 & 31.2 & 49.6 & 31.6 \\
    \bottomrule
    \end{tabular}
    \caption{System outputs compared to APE data \emph{postedits}. Results show that Knowledge Distillation (KD) on APE MT outputs also reduces the distance to APE postedits (higher BLEU, lower TER) . Baseline and KD systems are evaluated with a beam size of 5.}
    \label{tab:ape_pe_results}

\end{table*}

\subsection{Simulating Markings}

Error markings are simulated by comparing the hypothesis to the reference and marking
the longest common subsequence as correct, as proposed by \newcite{PetrushkovETAL:18}. We show an example of a data point in Table \ref{tab:datapoint_example}. Markings were extracted from the longest common subsequence calculations. For every token in the model hypothesis there is a corresponding reward. A reward is 0 when the corresponding token is not present in the reference and is 1 when the token was kept in the reference.

\subsection{Knowledge Distillation}

We want to showcase the advantage of our technique of enhancing supervised learning from human reference translations and from human postedits. In order to take advantage of the fact that human postedits indicate errors in machine translations instead of differences between machine translations and independent human references, we need to simulate the legacy machine translation system that produced the translations that were postedited. For this purpose we use APE data consisting of sources, MT outputs, and postedits. Since the legacy system is a black box to us, we carry out sequence-level knowledge distillation \cite{kim-rush-2016-sequence} on the machine translations provided in the train split of the APE dataset (cf.\ Section~\ref{sec:Data}). This allows us to emulate the legacy system by knowledge distillation and to consider the postedits in the APE dataset as feedback on the knowledge-distilled model. We present an overview of this process in Figure \ref{fig:postedits_and_kd}. 

As shown in Table \ref{tab:ape_mt_output_results}, after fine-tuning on the MT outputs in the train split of the APE data, we are able to produce translations that are more similar to the black-box systems than those of the pre-trained baseline system. 
Additionally, because the APE dataset's postedits were generated by correcting those MT outputs, Table \ref{tab:ape_pe_results} shows that the knowledge-distilled system's performance on the postedits is closer to the black-box system's performance than before distillation. 

\subsection{Online Learning}
\begin{figure*}[t!]
    \centering
    \includegraphics{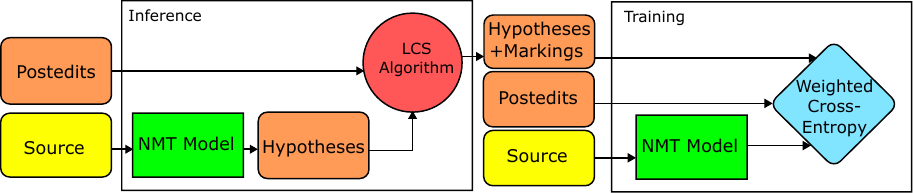}
    \caption{Once per epoch, we have our model run inference on all source sentences to generate hypothesis sentences. These then get compared to the postedits using the Longest Common Subsequence algorithm with tokens contained in the subsequence marked as good and those not in the subsequence marked as bad. Both the marked hypotheses and postedits are used as targets with a weighted cross-entropy loss function. The NMT model that generate the hypotheses and the model we train are the same model.}
    \label{fig:online_learning}

\end{figure*}

Our learning setup performs standard stochastic gradient descent learning on mini-batches. After every epoch, new system translations are produced and error markings are extracted by comparing the translations to references. This process is shown in Figure \ref{fig:online_learning}, showing that we produce error markings by comparing the model's output with the postedits and then use the marked hypotheses and the postedits to train the system. 

In preliminary experiments we found that computing error markings from a fixed initial set of system translations and using them as learning signals in iterative training appeared to bring initial improvements. Continued training, however, led to decreased performance. 
We conjecture that learning from constant marking signals can work for very small datasets (for example, \newcite{kreutzer-etal-2020-correct} used fewer than $1,000$ manually created markings for training), but it leads to divergence of parameter estimates on datasets that are one or two orders of magnitude larger, as in this work.


\begin{table*}[t]
    \centering
    \begin{tabular}{cccc}
    \toprule
        Dataset & Train & Dev & Test\\
        \midrule
        WMT17 (pre-train) & $5,919,142$ & & \\
        \midrule
        IWSLT14 (fine-tune) & $158,794$ & $7,216$ & $6,749$ \\
        WMT21 APE (fine-tune) & $7,000$ & $1,000$ & $1,000$ \\
        \bottomrule
    \end{tabular}
    \caption{Size of En-De datasets used for pre-training and fine-tuning: The WMT17 and IWSLT14 data consist of pairs of source and target sentences; the WMT21 APE data consists of triples of source, MT output, and postedited sentences.}
    \label{tab:dataset_sizes}
\end{table*}

\section{Data}
\label{sec:Data}


We use the WMT17 En-De dataset\footnote{\url{https://www.statmt.org/wmt17/translation-task.html}} for pre-training. 
Our data is pre-processed using the Moses tokenizer and punctuation normalization for both English and German implemented in Sacremoses\footnote{\url{https://github.com/alvations/sacremoses}}.

We first test our ideas on the IWSLT14 En-De dataset\footnote{\url{https://sites.google.com/site/iwsltevaluation2014/data-provided}} \cite{cettolo-etal-2012-wit3}. We download and pre-process the data using joey scripts\footnote{\url{https://github.com/joeynmt/joeynmt/blob/main/scripts/get_iwslt14_bpe.sh}}. The En-De dataset consists of transcribed TED talks and volunteer provided reference translations into the target languages. 

The APE dataset is from the WMT automatic postediting shared task 2021 \cite{akhbardeh-EtAl:2021:WMT}. The legacy system that produced the original MT outputs is based on a standard Transformer architecture \cite{vaswani-2017} and follows the implementation described by \newcite{ott-etal-2018-scaling}. This system was trained on publicly available MT datasets, including Paracrawl \cite{banon-etal-2020-paracrawl} and Europarl \cite{koehn-2005-europarl}, totalling 23.7M parallel sentences for English-German.
 The APE data consists of source, MT output, and postedit triples. The source data was selected from English Wikipedia articles. The MT outputs were provided by the legacy system and were postedited by professional translators. The sizes of the datasets are given in Table \ref{tab:dataset_sizes}.


\begin{table*}[t]
    \centering
    \begin{tabular}{cccl}
    \toprule
    System & References & Online markings  &         TER         \\
    \midrule
    a & 1.0 & 0.0 &       48.2        \\
    b & 0.9 & 0.1 & 48.1\\
    c & 0.7 & 0.3 & $48.0^{a,f}$ \\
    d & 0.5 & 0.5 & $47.8^{a,f}$ \\
    e & 0.3 & 0.7 & 48.3 \\
    \midrule
    f & $\emptyset$ & $\emptyset$ & 51.3 \\
    \bottomrule
    \end{tabular}
    \caption{Results from fine-tuning the WMT17 News model on out-of-domain IWSLT references. Numbers in the References and Online markings columns refer to interpolation weights given to that loss. The bottom row is the unchanged system, hence its interpolation values are $\emptyset$. The results show that, up to a threshold, increasing the weight given to Online markings improves TER scores. Superscripts denote statistically significant differences to indicated system at $p$-value $< 0.05$.}
    \label{tab:iwslt_results}
\end{table*}

\begin{table*}[t]
    \centering
    \begin{tabular}{cll}
    \toprule
    & System  & TER \\
    \midrule
    a & Baseline + Postedits & 31.3 \\
    b & Baseline + Postedits + Online Markings  & $30.7^a$ \\ 
    c & Baseline + KD + Postedits & 30.8 \\
    d & Baseline + KD + Postedits + Online Markings & $30.4^{ac}$ \\
    \bottomrule
    \end{tabular}
    \caption{Fine-tuned systems compared to WMT APE postedit test data. Results show that Online markings, when combined with learning from references, are able to improve our systems more than references alone. Even larger improvements are gained by systems trained by knowledge distillation (KD) on legacy translations. Interpolation weights are set to $0.5$. Superscripts indicate a significant improvement $p < 0.05$ over the indicated system.}
    \label{tab:wmt_ape_results}
\end{table*}

\section{Experiments}

\subsection{Experimental Setup}

We implement our loss function and data-loading on top of JoeyNMT \cite{kreutzer-etal-2019-joey}.\footnote{\url{https://github.com/joeynmt/joeynmt}} 
All that needs to be changed, in addition to adding weighting to the loss function, is a way of loading data and constructing combined batches such that each batch contains sources, hypotheses, weights, and postedits.
To do this, we duplicate each source twice in the batch and pair the first copy with the hypothesis and the second copy with the postedit. From the point of view of the model and loss function, the batch constructed for the combined objective does not differ from a normal batch with token-level weights. Batches constructed this way and in the usual manner can both contain the same number of tokens, but half of the target sequences in the combined batches come from the model's own translation of the training data.

Our baseline system is a standard Transformer model \cite{vaswani-2017}, pre-trained on WMT17 data for English-to-German translation \cite{bojar-EtAl:2017:WMT1}, and available through JoeyNMT\footnote{\url{https://www.cl.uni-heidelberg.de/statnlpgroup/joeynmt/wmt_ende_transformer.tar.gz}}. The model uses 6 layers in both the encoder and decoder with 8 attention heads each, and hyper-parameters as specified in the pre-trained JoeyNMT model's configuration file.

We compare the combined objective given in Equation (\ref{eq:combined-loss}) to standard supervised fine-tuning by continued training on references or postedits and to the pre-trained  model.

All systems share the same hyper-parameters except for the weighting of target tokens. The standard supervised learning method does not account for token-level weights and therefore all weights in the loss-function are set to $1$. For the contrastive marking method, we experimented with a range of interpolation values $\alpha$ on the IWSLT14 dataset to select the best value. The weighting of the tokens were set to $-0.5, 0.5$ in correspondence with the results from \newcite{kreutzer-etal-2020-correct}. 



\subsection{Experimental Results}

Since our work is concerned with learning from token-based feedback, we evaluate all systems according to Translation Edit Rate (TER) \cite{snover-etal-2006-study}. Furthermore, we provide the SacreBLEU \cite{post-2018-call} signatures\footnote{
TER: nrefs:1 $\vert$ ar:10000 $\vert$ seed:12345 $\vert$ case:lc $\vert$ tok:tercom $\vert$ norm:no $\vert$ punct:yes $\vert$ asian:no $\vert$ version:2.0.0} for evaluation configurations of evaluation metrics.
Statistical significance is tested using a paired approximate randomization test \cite{riezler-maxwell-2005-pitfalls}.

Table \ref{tab:iwslt_results} shows results from fine-tuning on independently created human references. A baseline model trained on WMT17 data (line f) is fine-tuned on references (line a) or on a combination of references and online markings (lines b-e, using different interpolation weights) from the TED talks domain. We see that up to a threshold, increasing the interpolation weight given to learning from online markings significantly improves TER scores up to $3.5$ points (line d) compared to the baseline (line f), and up to $0.5$ points compared to training from references only (line a). 

Table \ref{tab:wmt_ape_results} gives an experimental comparison of fine-tuning experiments on human postedits. A baseline model trained on news data is fine-tuned on postedit data from the Wikipedia domain. The postedit data is feedback on real MT outputs that we have trained on using knowledge distillation to emulate. Line a shows TER results for fine-tuning on postedits. This result can be improved significantly by $0.6$ TER by combined learning on postedits and online markings, using an interpolation weight of $0.5$ (line b). Lines c and d perform the same comparison of objectives for a model that has been trained via knowledge distillation (KD) of the legacy machine translations that were the input data for postediting. 
Comparing line d to line a, we see that by combined learning of a KD system on postedits and markings even larger gains, close to $1$ TER point, can be obtained. The improvements due to adding online markings are significant over training from postedits alone in all cases, and nominally, results for models adapted to the legacy machine translations via KD are better than for unchanged models trained on postedits.

An example showing the learning progress of the different approaches during the first epochs is given in Table \ref{tab:training_run_example}. The results of epoch 0 are given in the first block. It shows the system outputs of the models trained with knowledge distillation and the baselines before learning from postedits or markings. The KD models, given in lines c and d, already show better terminology translation (superstructure - Überbau, bases - Fundamente) than the baselines in lines a and b (superstructure - Superstruktur, bases - Stützpunkte). 
After one epoch, contrastive learning (lines b and d) and learning from postedits (lines a and c) correct "armored - gewagelt" and "armored - getrieben" to "armored - gepanzert", but only for KD models or if contrastive learning is used. Furthermore, constrastive learning of a KD model (line d) also corrects the translation  of "funnel" from "Funnels" to "Trichter".

 \begin{table*}[t!]
    \centering
    \small
    \begin{tabularx}{\linewidth}{l X}
    \toprule
        Source & the superstructure was armored to protect the bases of the turrets , the funnels and the ventilator ducts in what he termed a breastwork . \\
        \midrule
        Postedit & der Überbau wurde gepanzert , um die Fundamente der Türme , der Trichter und der Ventilatorkanäle in dem Bereich zu schützen , den er als Brustwehr bezeichnete . \\
    \bottomrule
    \toprule
        \multicolumn{2}{c}{Epoch 0} \\
        \midrule
        System & Hypothesis \\
         \midrule
        a & \hl{die Superstruktur} wurde ge\hl{trieben} , um die \hl{Stützpunkte} der \hl{Turm-} , der \hl{Funn ¤ rn-} und der Ventilat\hl{or die Herde} in dem , \hl{was} er als \hl{die} Brust\hl{st}be\hl{steigung bezeichnet hatte zu schützen} . \\
        b & \hl{die Superstruktur} wurde ge\hl{trieben} , um die \hl{Stützpunkte} der \hl{Turm-} , der \hl{Funn ¤ rn-} und der Ventilat\hl{or die Herde} in dem , \hl{was} er als \hl{die} Brust\hl{st}be\hl{steigung bezeichnet hatte zu schützen} . \\
        c & der Überbau wurde \hl{gewagelt} , um die Fundamente der Türme , \hl{die Funnels} und \hl{die} Ventilator\hl{en}kanäle in \hl{einem} Brust\hl{werk zu schützen} . \\
        d & der Überbau wurde \hl{gewagelt} , um die Fundamente der Türme , \hl{die Funnels} und \hl{die} Ventilator\hl{en}kanäle in \hl{einem} Brust\hl{werk zu schützen} . \\
    \bottomrule
    \toprule
        \multicolumn{2}{c}{Epoch 1} \\
        \midrule
        System & Hypothesis \\
         \midrule
        a & \hl{die Superstruktur} wurde ge\hl{zeichnet} , um die \hl{Stützen} der \hl{Turrets} , der \hl{Funnels} und der Ventilat\hl{or} in \hl{seiner Art} Brust\hl{work zu schützen} .\\
        b & \hl{die} Über\hl{bauung war} gepanzert , um die \hl{Grundstücke} der \hl{Turrets} , der \hl{Funnels} und der V\hl{aterfun}kan\hl{ten} in dem , \hl{was} er als Brust\hl{werk nannte , zu schützen} .\\
        c & der \hl{Super}bau wurde gepanzert , um die \hl{Stützpunkte} der \hl{Turrets} , der \hl{Funnels} und der Ventilator\hl{entötungen} in \hl{einer so genannten} Brust\hl{arbeit zu schützen} . \\
        d & der Überbau wurde gepanzert , um die Fundamente der Türme , der Trichter und der Ventilator\hl{kan}kanäle zu schützen , \hl{was} er als Brust\hl{werk nannte} . \\
    \bottomrule
    \end{tabularx}
    \caption{Here we show the beginning of a training trajectory for a single example from the APE dataset. Above is the source and the postedit from the dataset, after which follows the first three epochs. Because translations and markings are generated before the beginning of an epoch, epoch 0 contains outputs from the knowledge distilled (KD) (lines c and d) and baseline systems (lines a and b). The systems letters correspond to those in Table \ref{tab:wmt_ape_results}, indicating learning from postedits in lines a and c, and learning additionally from the contrastive markings in lines b and d.  Models c and d have seen the MT side of this dataset beforehand and are already more capable of translating terminology such as "superstructure" to "Überbau". After one epoch, we see that the KD models and the contrastive learning objective models are able to correct "gewagelt" and "getrieben" to "gepanzert" as the translation of "armored". Because we use subword tokens, we have markings on portions of words. Although "Überbau" is a part of "Überbauung", the subwords used to construct them differ, leading to "bau" in "Überbauung" being marked as incorrect.
    }
    \label{tab:training_run_example}
\end{table*}

\section{Discussion}




Our experimental results in Table \ref{tab:wmt_ape_results} show that online markings combined with references or postedits bring greater improvements than supervised learning on references or postedits alone, and moreover, the knowledge distilled models benefit more from the provided feedback. This suggests that the more related the feedback is to the system's own output, the more can be learned from the feedback. 

Furthermore, this result has implications for how to best use postedits. Postedits are often treated as new reference translations for the sources and used to train new systems, whereas the original MT outputs are discarded. However, fine-tuning the original system  on the postedits may yield larger improvements than training a new, unrelated model on the source and postedit alone.

Lastly, we believe that our results can be interpreted as the effect of mitigating exposure bias. The pre-trained model is exposed not only to reference translations, but to its own trajectories. Even if the model's trajectory is far from the gold reference and multiple tokens in its history are incorrect, it will be rewarded if it predicts a token that is in the output. This may enable it to return to a more rewarding trajectory.

\section{Conclusion}

In this work we present a way to combine postedits and word-level error markings extracted from the edit operations between the postedit and the MT output to learn more than what the postedit alone is able to provide. Experimentally, we try this on systems unrelated to the legacy system, whose outputs were originally postedited, and on a simulation of the legacy system we create via knowledge distillation. We show that these contrastive markings are able to bring significant improvements to TER scores and we hypothesize this is because they are able to target insertion errors that contribute to higher TER scores. Additionally, learning from the model's own output may allow it to learn how to correct itself after making an error if it is later rewarded for correct outputs.



\bibliographystyle{eamt23}
\bibliography{eamt23}
\end{document}